\documentclass[letterpaper, 10 pt, conference]{ieeeconf}  

\IEEEoverridecommandlockouts                              

\usepackage{amsmath} 
\usepackage{amssymb}  
\usepackage[font=small,skip=2pt]{caption}
\usepackage{graphicx}
\usepackage{capt-of} 
\usepackage{overpic} 
\usepackage[export]{adjustbox} 
\usepackage{minibox} 
\usepackage{textcomp} 
\usepackage{tabularx}
\usepackage{multirow}
\usepackage{booktabs}
\usepackage{arydshln}
\usepackage{adjustbox} 
\usepackage{pgfplots} 
\usepackage{siunitx}
\usepackage{pifont}
\usepackage{stfloats}
\usepackage{makecell}
\usepackage{wrapfig}
\usepackage{url}
\usepackage{nicematrix}  

\usepackage{algpseudocode,algorithm,algorithmicx}

\algrenewcommand\algorithmicrequire{\textbf{Precondition:}}
\algrenewcommand\algorithmicensure{\textbf{Postcondition:}}
\newcommand{\cmark}{\ding{51}}%

\usepackage[caption=false, font=scriptsize]{subfig} 

\usepackage{cite}

\DeclareSIUnit{\million}{\text{Mio.}}

\makeatletter
\let\NAT@parse\undefined
\makeatother

\usepackage[colorlinks=false,linkcolor=blue,citecolor=blue]{hyperref}

\title{\LARGE \bf
TempBEV: Improving Learned BEV Encoders with\\
Combined Image and BEV Space Temporal Aggregation
}

\author{Thomas Monninger$^{1,2}$, Vandana Dokkadi$^{3,*}$, Md Zafar Anwar$^{4,*}$, Steffen Staab$^{2,5}$
\thanks{$^{1}$Mercedes-Benz Research \& Development North America, Sunnyvale, CA, USA (email: thomas.monninger@mercedes-benz.com)}%
\thanks{$^{2}$University of Stuttgart, Institute for Artificial Intelligence, Stuttgart, Germany (email: steffen.staab@ki.uni-stuttgart.de)}%
\thanks{$^{3}$University of Massachusetts Amherst, Amherst, MA, USA}%
\thanks{$^{4}$Pennsylvania State University, University Park, PA, USA}%
\thanks{$^{5}$University of Southampton, Southampton, United Kingdom}%
\thanks{$^{*}$Work was done during an internship at Mercedes-Benz Research \& Development North America.}%
}

\newcommand\copyrighttext{\footnotesize \textcopyright~2024 IEEE. Personal use of this material is permitted.  Permission from IEEE must be obtained for all other uses, in any current or future media, including reprinting/republishing this material for advertising or promotional purposes, creating new collective works, for resale or redistribution to servers or lists, or reuse of any copyrighted component of this work in other works.
}%

\newcommand\copyrightnotice{%
    \begin{tikzpicture}[remember picture,overlay]%
     \node[anchor=south, xshift=0pt, yshift=12pt] at (current page.south)%
     {\fbox{\parbox{\dimexpr\textwidth-\fboxsep-\fboxrule\relax}{\copyrighttext}}};%
     \end{tikzpicture}%
}

\begin{document}
\maketitle
\thispagestyle{empty}
\pagestyle{empty}

\newcommand{\etal}{\textit{et al}.}
\newcommand{\ie}{\textit{i}.\textit{e}.}
\newcommand{\eg}{\textit{e}.\textit{g}.}
\newcommand{\vs}{\textit{vs}. }

\begin{abstract}
Autonomous driving requires an accurate representation of the environment. 
A strategy toward high accuracy is to fuse data from several sensors.
Learned Bird's-Eye View (BEV) encoders can achieve this by mapping data from individual sensors into one joint latent space.
For cost-efficient camera-only systems, this provides an effective mechanism to fuse data from multiple cameras with different views.
Accuracy can further be improved by aggregating sensor information over time.
This is especially important in monocular camera systems to account for the lack of explicit depth and velocity measurements, such that decision-critical information about distance distances and motions of other objects is easily accessible.
Thereby, the effectiveness of developed BEV encoders crucially depends on the operators used to aggregate temporal information and on the used latent representation spaces.

We analyze BEV encoders proposed in the literature and compare their effectiveness, quantifying the effects of aggregation operators and latent representations. 
While most existing approaches aggregate temporal information \emph{either} in image \emph{or} in BEV latent space, our analyses and performance comparisons suggest that these latent representations exhibit complementary strengths.
Therefore, we develop a novel temporal BEV encoder, TempBEV, which integrates aggregated temporal information from both latent spaces. 
We consider subsequent image frames as stereo through time and leverage methods from optical flow estimation for temporal stereo encoding.

Empirical evaluation on the NuScenes dataset shows a significant improvement by TempBEV over the baseline for 3D object detection and BEV segmentation. 
The ablation uncovers a strong synergy of joint temporal aggregation in the image and BEV latent space.
These results indicate the overall effectiveness of our approach and make a strong case for aggregating temporal information in both image and BEV latent spaces.

\end{abstract}


\section{Introduction}

Autonomous driving requires a perception system that can derive an accurate representation of the environment, which includes both  dynamic objects and static elements.
Perception tasks include 3D object detection \cite{chen2016monocular} (estimating bounding boxes for dynamic objects), Bird's-Eye View (BEV) segmentation \cite{pan2020cross} (performing pixel-level semantic classification in 2D top-down view), motion tracking, etc.
Autonomous vehicles depend on the outputs of these tasks to understand the structure of their environment and make informed decisions.
While monocular multi-camera systems offer a cost-effective solution, they provide only projected views and lack explicit depth information \cite{lu2021geometry}.
This poses challenges for tasks inherently requiring 3D perception.
{\copyrightnotice}

One way of improving depth estimation is temporal aggregation, a process of combining sensor information over multiple time steps, for example through temporal stereo \cite{davis2003spacetime, wang2022dfm}.
Temporal aggregation also allows for mitigating occlusions, reducing measurement uncertainty, and estimating motions of other dynamic objects by considering several views from the same cameras at different times, reaching a comprehensive representation of the traffic scene \cite{monninger2023scene}.

Temporal aggregation of camera frames requires determining the corresponding features across subsequent frames.
The camera pose changes due to the ego-motion of the vehicle \cite{huang2022bevdet4d}, on which the cameras are mounted, and other dynamic objects \cite{wang2023streampetr} move in the 3D environment affecting their position in the different frames.
Image space represents a 2D projection of the 3D environment, making linear motions of dynamic objects appear non-linear.
Moreover, objects in image space might also change appearance over time, for example through changes in perspective, lighting, or occlusion, adding to the complexity of finding correspondences.

\begin{figure}[tpb]
	\centering
 \vspace{0.6em}
	\includegraphics[width=\linewidth]{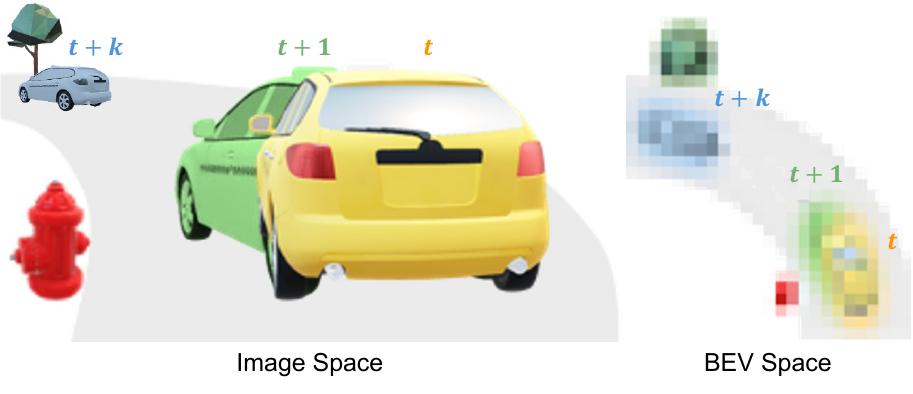}
    \vspace{-1.2em}
	\caption{Exemplary traffic scene at three time steps represented in high-resolution Image \vs low-resolution BEV space. Same car is colored differently at different time steps only for better visibility.}
    \vspace{-0.8em}
	\label{fig:image_vs_BEV}
 \end{figure}

Most recent works \cite{li2023bevdepth, huang2022bevdet4d, wang2023streampetr,li2022bevformer, yang2023bevformerv2, zhang2022beverse, qing2023dort} perform temporal aggregation in BEV space.
BEV space is an orthogonal projection of the scene into a 2D top-down view, preserving a linear appearance of linear motions.
Lifting is performed to transform the camera frames from projected image space into 3D space, followed by an orthogonal projection to BEV space.
Thereby, BEV space representation reduces the complexity of finding correspondences.
However, the uncertainty of depth estimation  \cite{lu2021geometry} as well as the lower resolution of BEV space \cite{park2022solofusion} affect the estimate of object positions in BEV representation.
This is illustrated in Fig. \ref{fig:image_vs_BEV}, which shows a traffic scene in image \vs BEV representation.
Aggregating only in BEV space misses features present in high-resolution image space, indicating the complementary nature of the image and BEV space.

We argue that it is beneficial to consider both image and BEV space for temporal aggregation to combine the different representational strengths.
To design a suitable model, we survey existing literature and perform a comparative study of different aggregation mechanisms including attention, convolution, and max pooling.
With the insights gained from this study, we propose a novel \textbf{Temp}oral \textbf{BEV} encoder, \textbf{TempBEV}, that effectively combines temporal aggregation in image and BEV space.
In summary, the main contributions of this paper are:
\begin{itemize}
\item We provide a survey on temporal aggregation mechanisms for learned BEV encoders.
\item We compare learned temporal aggregation operators through systematic experiments and investigate their effect on object detection and BEV segmentation tasks. 
\item We propose TempBEV to combine temporal aggregation in image and BEV space and investigate its effectiveness through experimental evaluation. 
\item We apply methods from optical flow for image space aggregation and show the benefit of transfer learning.
\end{itemize}

\section{Survey on Temporal Aggregation}\label{sec:survey_temporal_aggregation}
We  survey temporal aggregation mechanisms in recent BEV models (cf.\  Table \ref{tab:temp_bev_survey}).
Models are distinguished based on the following five key aspects: aggregation operator, recurrent or parallel aggregation and number of steps, feature space, motions, and learning tasks.

{\begin{table*}[tb]
    \vspace{0.4em}
	\caption{Survey of learning-based temporal BEV aggregation methods on 3D object detection (Det) and BEV segmentation (Seg).}
	\label{tab:temp_bev_survey}
	\centering
	\resizebox{\textwidth}{!}{
	\begin{NiceTabular}{l|lccc|cc|c|ccc}

		 \toprule
   
		 \multicolumn{1}{l}{\textbf{Methods}}   &   \multicolumn{4}{c}{\textbf{Aggregation}}&  \multicolumn{2}{c}{\textbf{Feature Space}}&\multicolumn{1}{c}{\textbf{Motions}} & \multicolumn{3}{c}{\textbf{Learning Tasks}}\\
               
         Publication                              & Operator                &Recurrent&Parallel&\# Steps     &   Image  &  BEV   & Object-based & Det& Seg   & Other  \\ 
         \midrule
         BEVDepth \cite{li2023bevdepth}           &  Convolution            &        &  \cmark &2            &          & \cmark &         & \cmark &        &        \\
         BEVDet4D \cite{huang2022bevdet4d}        &  Convolution            &        &  \cmark &2            &          & \cmark &         & \cmark &        &        \\
         BEVFormer \cite{li2022bevformer}         &  Deformable Attention   & \cmark &         &4            &          & \cmark &         & \cmark & \cmark &        \\
         BEVFormer v2 \cite{yang2023bevformerv2}  &  Convolution            &        &  \cmark &4            &          & \cmark &         & \cmark &        &        \\
         BEVStereo \cite{li2023bevstereo}         &  Convolution            &        &  \cmark &2            &  \cmark  & \cmark &         & \cmark &        &        \\
         BEVStitch \cite{can2022bevstitch}        &  Max pooling            &        &  \cmark &4            &  \cmark  &        &         &        & \cmark &        \\
         BEVerse \cite{zhang2022beverse}          &  Convolution            &        &  \cmark &3            &          & \cmark &         & \cmark & \cmark & \cmark \\
         DORT \cite{qing2023dort}                 &  Convolution            & \cmark &         &16           &          & \cmark & \cmark  & \cmark &        & \cmark \\
         DfM \cite{wang2022dfm}                   &  Convolution            &        &  \cmark &2            &  \cmark  &        &         & \cmark &        &        \\
         DynamicBEV \cite{yao2023dynamicbev}      &  Attention              & \cmark &         &8            &          & \cmark & \cmark  & \cmark &        &        \\
         FIERY \cite{hu2021fiery}                 &  Attention+Convolution  &        &  \cmark &3            &          & \cmark &         &        & \cmark & \cmark \\
         Fast-BEV \cite{huang2023fast,li2023fast} &  Convolution            &        &  \cmark &4            &          & \cmark &         &        &        &        \\
         HoP \cite{zong2023hop}                   &  Deformable Attention   & \cmark &  \cmark & 4 / 2        &          & \cmark &         & \cmark &        &        \\
         Img2Maps \cite{saha2022imgtomap}         &  Axial Attention        &        &  \cmark &4            &          & \cmark &         &        & \cmark &        \\
         MaGNet \cite{bae2022multi}               &  Convolution            &        &  \cmark &5            &  \cmark  &        &         &        &        & \cmark \\
         MVSNet \cite{yao2018mvsnet}              &  Convolution            &        &  \cmark &3            &  \cmark  &        &         &        &        & \cmark \\
         OCBEV \cite{qi2023ocbev}                 &  Deformable Attention   & \cmark &         &4            &          & \cmark &         & \cmark &        &        \\
         PETRv2 \cite{liu2023petrv2}              &  Attention              &        &  \cmark &2            &  \cmark  &        &         & \cmark & \cmark & \cmark \\
         PointBEV \cite{chambon2023pointbev}      &  Sub-manifold Attention &        &         &8            &          & \cmark &         &        & \cmark &        \\
         PolarDETR \cite{chen2022polardetr}       &  Attention              &        &  \cmark &2            &  \cmark  &        &         & \cmark &        & \cmark \\
         PolarFormer \cite{jiang2023polarformer}  &  Attention              &        &  \cmark &2            &          & \cmark &         & \cmark &        &        \\
         ST-P3 \cite{hu2022stp3}                  &  Attention+Convolution  &        &  \cmark &3            &          & \cmark & \cmark  &        & \cmark & \cmark \\
         STS \cite{wang2022sts}                   &  Group-wise correlation &        &  \cmark &2            &  \cmark  &        &         & \cmark &        &        \\
         SoloFusion \cite{park2022solofusion}     &  Convolution            &        &  \cmark &17           &  \cmark  & \cmark &         & \cmark &        &        \\
         SparseBEV \cite{liu2023sparsebev}        &  Attention              &        &  \cmark &8            &          & \cmark & \cmark  & \cmark &        &        \\
         StreamPETR \cite{wang2023streampetr}     &  Attention              & \cmark &         &4            &          & \cmark & \cmark  & \cmark &        & \cmark \\
         TBP-Former \cite{fang2023tbp}            &  Deformable Attention   &        &  \cmark &3            &          & \cmark &         &        & \cmark & \cmark \\
         UVTR \cite{li2022unifying}               &  Convolution            &        &  \cmark &5            &          & \cmark &         & \cmark &        &        \\
         UniFusion \cite{qin2023unifusion}        &  Deformable Attention   &        &  \cmark &7            &          & \cmark &         &        & \cmark &        \\
         VideoBEV \cite{han2023videobev}          &  Convolution            & \cmark &         &8            &          & \cmark &         & \cmark & \cmark & \cmark \\
            
         \bottomrule
		
    \end{NiceTabular}
    }
    \vspace{-1.0em}
\end{table*}}

\subsection{Aggregation Operator}
The aggregation operator defines the mathematical operation that is used to combine information from multiple time steps.
Commonly used temporal aggregation operators are attention, convolution, and max pooling.
Most approaches use the attention mechanism, owing to its effectiveness and expressiveness.
More specifically, deformable self-attention \cite{zhu2021deformable} is frequently employed to address computational complexity for real-time applications.
A simple alternative is max pooling, which requires no additional parameters (\eg, BEVStitch \cite{can2022bevstitch}).
A few other approaches exist, such as STS \cite{wang2022sts}, which employs group-wise correlations processed by Multi-Layer Perceptrons.

\subsection{Recurrent or Parallel Aggregation}

For a given time step $t$, let $U_{t}=\{u^1_{t}, \ldots, u^n_{t}\}$ be the set of $n$ inputs (\eg, corresponding to the images from $n$ different cameras).
A key choice is whether $U_{t-k:t}$ is aggregated in a recurrent or parallel manner to combine features into a latent state $X$.
Recurrent aggregation conditions the current state $X_{t}$ on $X_{t-1}$.
This enables information flow over a longer time horizon, primarily limited by the capacity of the latent state.
All works in this survey that uses recurrent aggregation  (\eg, \cite{li2022bevformer}) perform it in BEV space with $X$ being a BEV grid.
Mathematically, at time \textit{t}, the latent state $X_{t}$ can be expressed as a function $f_{recurrent}$  of the input $U_{t}$ and the previous latent state $X_{t-1}$:
\begin{equation}
	{X}_{t} = f_{\operatorname{recurrent}} \left( {U}_{t}, {X}_{t-1} \right)
\end{equation}
Alternatively, parallel aggregation directly combines input from a fixed number of time steps (\eg, \cite{yang2023bevformerv2}).
This approach can learn individual aggregations specific for each time step.
However, computation grows linearly, limiting the number of time steps.
This effectively constrains the available temporal information to a limited time horizon.
Mathematically, $X_{t}$ can be expressed as a function $f_{parallel}$ of the inputs from the last \textit{k} time steps.
\begin{equation}
	{X}_{t} = f_{\operatorname{parallel}} \left( {U}_{t-k:t} \right)
\end{equation}
The survey indicates that both parallel and recurrent aggregation techniques are prevalent, with no clear advantage of one over another.
To the best of our knowledge, HoP \cite{zong2023hop} is the only model using both parallel and recurrent aggregation.
Their setup performs recurrent aggregation of the BEV grids and also adds a parallel aggregation module to process two BEV grids directly.
Furthermore, it is important to specify the number of time steps considered in each approach.
For parallel aggregation, this refers to the number of aggregated time steps.
For recurrent aggregation, it denotes the extent to which the recurrent computation graph unfolds during training, while the inference setting is not limited to a specific number of time steps. 
Several studies have investigated the impact of the number of time steps, yielding diverse conclusions.
BEVFormer \cite{li2022bevformer} employs 4 time steps (equivalent to \SI{2}{\second}), VideoBEV \cite{han2023videobev} and StreamPETR \cite{wang2023streampetr} utilize 8 time steps (\SI{4}{\second}), and SOLOFusion \cite{park2022solofusion} extends this further to 17 time steps (\SI{8.5}{\second}).
It is worth noting that these studies do not adequately analyze the potential of long time horizons for detecting static elements in the scene.
They primarily focus on object detection, where a quick performance saturation is expected due to the dynamic nature of objects.

\subsection{Feature Space}\label{sec:feature_space}
Related works mainly use representation in two feature spaces: image and BEV (see Fig. \ref{fig:image_vs_BEV}).
Image feature space is derived from camera images that contain information in projected camera views.
BEV feature space is a joint latent space where features from multiple sensor views are mapped into and from which BEV and 3D representations are decoded, imposing spatial constraints \cite{harley2023simplebev}.
This concept was pioneered by Philion and Fidler \cite{philion2020lift}, where they lift each image to a 3D volume and then map all volumes into a joint BEV grid.
Typically learned BEV spaces are of low resolution due to computational constraints, as illustrated in Fig. \ref{fig:image_vs_BEV}.

Most related works aggregate information in BEV space.
Representation in BEV space is affected by the uncertainty in depth estimation that is propagated through the lifting step \cite{lu2021geometry}, but also preserves a linear appearance of linear motions, simplifying the aggregation process.
Only a few models, such as PETRv2 \cite{liu2023petrv2}, STS \cite{wang2022sts}, DfM \cite{wang2022dfm}, BEVStitch \cite{can2022bevstitch}, and PolarDETR \cite{chen2022polardetr} conduct aggregation purely in image feature space.
These models typically leverage methods for learning-based temporal stereo.
To the best of our knowledge, SoloFusion\cite{park2022solofusion} and BEVStereo\cite{li2023bevstereo} are the only approaches that perform temporal aggregation in both feature spaces.
BEVStereo uses depth estimation from two consecutive frames to improve the quality of the lifting and performs parallel temporal aggregation in BEV space.
SoloFusion employs parallel aggregation in image space and BEV space.
A high-resolution cost volume is created from image space to perform stereo matching between current and previous camera frame.
This is complemented by a low-resolution aggregation in BEV space over a longer time horizon.
Both approaches have a fixed time horizon by performing parallel temporal aggregation.

\subsection{Motions}\label{sec:motion_compensation}
Sensors equipped on an autonomous vehicle observe a superposition of two kinds of motions: ego-motion \cite{huang2022bevdet4d}, which alters the reference frame for all sensors on the vehicle, and the dynamic motion of objects \cite{wang2023streampetr, qing2023dort} in the environment.

Ego-motion refers to the movement of the ego vehicle over time relative to static elements in the scene.
This concept entails that all sensors mounted on the vehicle undergo the same transformation. 
In BEV space, ego-motion is compensated by applying the inverse of the rotation and translation to transform previous measurements into the current frame of reference. 
This process ensures that static elements are consistently represented at the same spatial locations across different time steps, thereby facilitating the aggregation.
BEVDet4D \cite{huang2022bevdet4d} offers deeper insights into the benefits of ego-motion compensation to the learning process.
All works in this survey that perform temporal aggregation in BEV space apply ego-motion compensation to $X_{t-1}$. 
Ego-motion compensation in image space is a more challenging problem that only some works address \cite{wang2022sts}.

The second kind of motion in the scene is the motion of other objects.
For the motion of other objects, few approaches \cite{wang2023streampetr, liu2023sparsebev, hu2022stp3, qing2023dort} use an explicit representation of dynamic objects in the latent state $X$.
The latent state of the model is constrained to capture dynamic objects and their motions effectively.
In Table \ref{tab:temp_bev_survey} this is referred to as object-based motion.
A notable example is StreamPETR \cite{wang2023streampetr}, which limits the latent space $X_t$ to cover only sparse object queries from previous time steps, guiding the model to focus on dynamic objects in the scene.

\subsection{Learning Tasks}
The learning tasks encompass object detection \cite{chen2016monocular}, BEV segmentation \cite{pan2020cross}, and other miscellaneous objectives. 
Object detection involves estimating the 3D bounding boxes of objects within the scene and is a fundamental task addressed by almost all approaches. 
BEV segmentation, on the other hand, focuses on representing static elements of the environment, and only some approaches include this task in their scope. 
Integrating other tasks into an end-to-end learning setup is covered by only a few works so far.
The "other" category may involve predicting trajectories \cite{zhang2022beverse, hu2021fiery, fang2023tbp}, object tracking \cite{han2023videobev, wang2023streampetr,qing2023dort}, occupancy maps \cite{hu2022stp3}, or other downstream tasks.

\section{Approach} \label{sec:approach}

\subsection{Problem Statement} \label{sec:problem}
For a given time step $t$, let $U_{t}=\{u^1_{t}, \ldots, u^n_{t}\}$ be the set of image frames from the $n$ monocular cameras mounted on the ego vehicle, $e_{t}$ be the ego-motion vectors representing translation and rotation from $t-1$ to $t$,
$B_{t}$ be the set of 6D vectors representing 3D location and dimension of bounding boxes of dynamic objects visible from the ego vehicle,
and $S_{t}$ be the segmentation in BEV space as a 2D grid representing the ground plane with the origin at the ego-vehicle, where each grid pixel represents the class of the static element at that location.
The goal is to find a function $h$ that returns bounding boxes $B_{t}$ and BEV segmentation $S_{t}$ for a given sequence of sets of  image frames ${U}_{t-k:t}$:
\begin{equation}
    \left( B_{t}, S_{t} \right) = h \left( U_{t-k:t},\ e_{t-k:t} \right).
    \label{eq:problem_statement}
\end{equation}

Function $h$ is typically realized in an encoder-decoder fashion where the encoder $f$ performs temporal aggregation and $g$ is the task-specific decoder: $h\left(x\right) = g\left(f(x)\right)$.
The objective of our study is to propose a temporal encoder $f$ that effectively extracts temporal features that can be used by $g$ to estimate $B_{t}$ and $S_{t}$.

\subsection{Approach for Temporal Aggregation} \label{sec:approach_temporal_aggregation}

Temporal aggregation is the process of combining inputs ${U}_{t-k:t}$ to improve the prediction of $B_{t}$ and $S_{t}$ at time $t$.
Our survey shows that previous research has predominantly focused on temporal aggregation either in BEV space or image space (see Sec. \ref{sec:feature_space}).
Eq. \ref{eq:aggregation_in_BEV} formalizes aggregation in BEV space, Eq. \ref{eq:aggregation_in_Image} in image space, where $\operatorname{lift}$  is the operation that encodes and lifts projected camera features into 3D space, followed by projection into BEV space.
\begin{equation}
    \left(B_{t}, S_{t} \right) = g\left( f_{\operatorname{BEV}} \left( \operatorname{lift}\left({U_{t-k:t}}\right) \right) \right)
    \label{eq:aggregation_in_BEV}
\end{equation}
\begin{equation}
    \left(B_{t}, S_{t} \right) = g\left(\operatorname{lift}\left( f_{\operatorname{img}} \left( {U_{t-k:t}} \right) \right) \right)
    \label{eq:aggregation_in_Image}
\end{equation}
We analyze the pros and cons of temporal aggregation in image \vs BEV space using Fig. \ref{fig:image_vs_BEV}.
Consecutive frames in image space offer precise cues of motion due to high resolution and low uncertainty, as visible in Fig. \ref{fig:image_vs_BEV} for time $t$ and $t+1$.
Those visual cues are desirable for detecting the motion of dynamic objects over short time horizons.
Image space aggregation over long time horizons is more challenging since ego-motion compensation cannot be directly applied to features in image space \cite{wang2022sts}.
This can make even static elements appear increasingly different over long time horizons, increasing the difficulty of finding correspondences for aggregation, as visible in Fig. \ref{fig:image_vs_BEV} for time $t$ and $t+k$.

Conversely, changes across short time horizons are less apparent in BEV space due to lower resolution and higher uncertainty induced by lifting \cite{lu2021geometry}, as visible in BEV space in Fig. \ref{fig:image_vs_BEV}.
However, representing features in BEV space allows for ego-motion compensation, which keeps static elements at the same location in the grid.
This enables aggregation over longer time horizons, where a broader range of observations must be aggregated from diverse perspectives and varying occlusions.
Our understanding of the time horizon is consistent with the work in SOLOFusion \cite{park2022solofusion}, where they demonstrate that temporal aggregation over short and long time horizons is complementary. 

Based on our analysis, we hypothesize that aggregation in image and BEV space is complementary and that combining them leverages the strengths of both representations.
To this end, we propose a model that effectively combines temporal aggregation in image and BEV space by extracting temporal features from both representations as formalized in Eq. \ref{eq:aggregation_in_both_Image_and_BEV}.
\begin{equation}
        \left( B_{t}, S_{t} \right) = g \left( f_{\operatorname{BEV}} \left( \operatorname{lift} \left( {U_{t-k:t}}, f_{\operatorname{img}} \left( {U_{t-k:t}} \right) \right) \right) \right)
    \label{eq:aggregation_in_both_Image_and_BEV}
\end{equation}

\subsection{TempBEV Model} \label{sec:tempbev}
We design a novel model TempBEV.
As per common practice \cite{zong2023hop, yuan2024streammapnet, hu2023uniad}, we use BEVFormer \cite{li2022bevformer} as the starting point of our implementation.
Eq. \ref{eq:aggregation_tempbev} formalizes the temporal aggregation of TempBEV, refining Eq. \ref{eq:aggregation_in_both_Image_and_BEV}.
For image space aggregation $f_{\operatorname{img}}$, we add a temporal stereo encoder \cite{laga2020survey} that performs parallel aggregation of $U_{t-1:t}$.
For BEV space aggregation, we keep the recurrent mechanism of BEVFormer to cover a long time horizon, so $f_{\operatorname{BEV}}$ uses $X_{t-1}$.
In addition, $f_{\operatorname{BEV}}$ also integrates the lifted encodings from $U_t$ and $f_{\operatorname{img}}$.
\begin{equation}
        \left( B_{t}, S_{t} \right) = g \left( f_{\operatorname{BEV}} \left( \operatorname{lift} \left( {U_{t}}, f_{\operatorname{img}} \left( {U_{t-1:t}} \right) \right) , X_{t-1} \right) \right)
    \label{eq:aggregation_tempbev}
\end{equation}

By leveraging both image and BEV feature space for temporal aggregation, TempBEV can learn from data which information to aggregate in BEV space and what temporal features to extract directly from the image space. 

\begin{figure*}[tpb]
	\centering
  \vspace{0.4em} 
	\includegraphics[width=\textwidth,trim={0 0 0 0.0cm},clip]{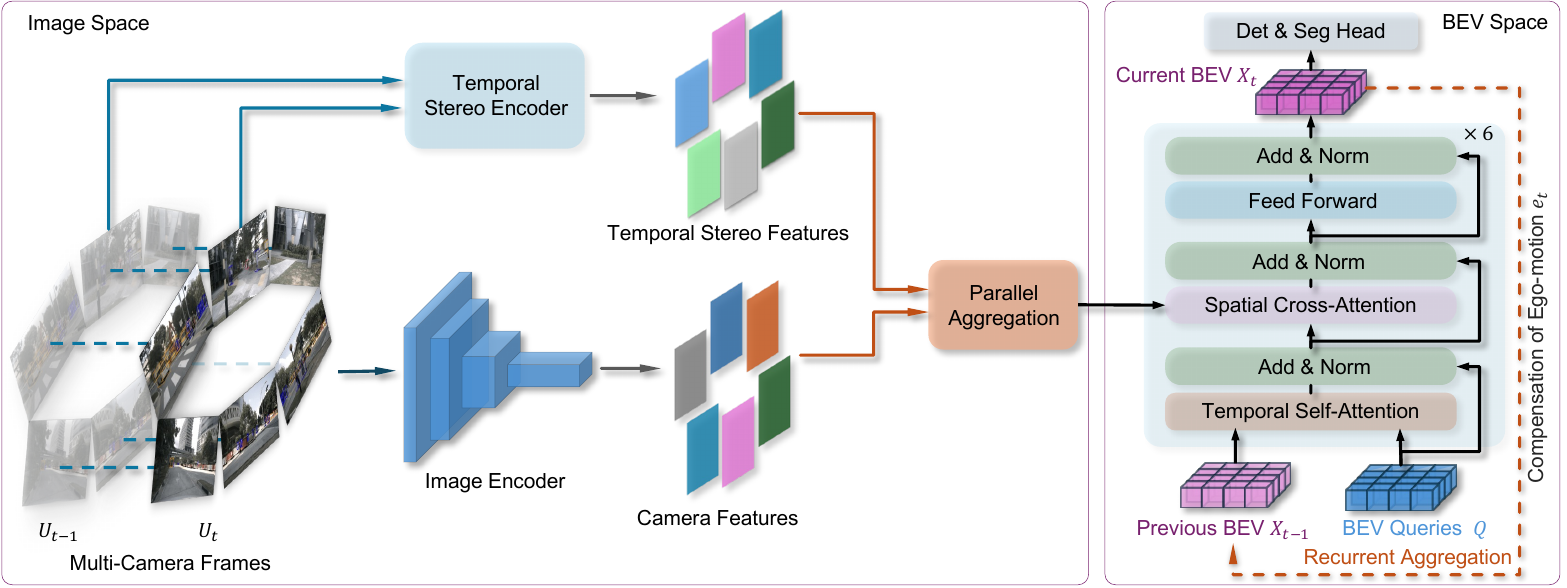}
	\caption{Proposed TempBEV model architecture with image space and BEV space temporal aggregation mechanisms colored orange.}
\vspace{-0.7em}	
 \label{fig:tempbev_architecture}
\end{figure*}

Figure \ref{fig:tempbev_architecture} is inspired by BEVFormer \cite{li2022bevformer} and provides a visualization of the architecture, illustrating the temporal aggregation mechanisms employed by TempBEV. 
In image space, a parallel aggregation module combines camera features encoded by an image encoder and temporal stereo features encoded by a temporal stereo encoder.
In BEV space, a transformer encoder layer performs temporal self-attention and spatial cross-attention.
To quantify the effect of temporal stereo encoding on the detection of dynamic objects and static elements separately, we perform 3D object detection and map BEV segmentation using task-specific heads.
Details on the aggregation mechanisms are explained below:

\subsubsection{TempBEV Aggregation in Image Space}
Camera frames from multiple views $U_t$ are individually encoded with a shared image encoder to generate camera features. 
In addition, a temporal stereo encoder $f_{\operatorname{img}}$ extracts temporal features from individual image pairs $u^k_{t-1}$ and $u^k_t$ with $k=\{1,\dots,n\}$.
The temporal stereo encoder draws inspiration from optical flow approaches.
The intuition is that motions in the scene induce optical flow, so extracting features for optical flow from image space helps capture these motions.
To confirm our intuition, we select the simplest model for learned optical flow, FlowNet \cite{dosovitskiy2015flownet}, and use its encoder as the temporal stereo encoder in our TempBEV architecture. 
The FlowNet encoder consists of 10 convolution layers, each followed by Batch Normalization and Leaky ReLU activation.
The temporal stereo encoder is shared between all cameras to minimize model size. 

Image encoder and temporal stereo encoder both provide 4 outputs of intermediate layers.
A parallel aggregation module combines temporal stereo features and camera features on the 4 different intermediate resolutions using a 2-layer CNN each.
This two-step parallel aggregation captures temporal features over a short time horizon in image feature space. 

\subsubsection{TempBEV Aggregation in BEV Space}
In BEV space, a transformer encoder layer performs temporal self-attention with the previous BEV grid $X_{t-1}$ in a recurrent fashion.
Furthermore, spatial cross-attention is used to lift the aggregation of camera features extracted from $U_t$ and temporal stereo features extracted by $f_{\operatorname{img}}$ into BEV space.
Learned BEV queries $Q$ are iteratively refined by applying 6 instances of this transformer encoder layer before passing it to the task-specific heads.
This setup creates a recurrent chain that facilitates the propagation of information across multiple time steps, enabling the capture of temporal features over long time horizons.

\section{Experiments} \label{sec:experiments}

\subsection{Dataset and Evaluation Metrics}\label{sec:metrics}
We use the NuScenes dataset \cite{caesar2020nuscenes} for our experiments, which provides data points at \SI{2}{\hertz}.
Those include images from 6 monocular cameras $U_{t}$ and corresponding ground truth bounding boxes $B_{t}$ and BEV segmentation $S_{t}$ for static elements including road surface ("Road"), dividers ("Lane"), and pedestrian crossings ("Cross") .
The performance on 3D object detection is evaluated using NuScenes Detection Score (NDS) \cite{caesar2020nuscenes} and mean Average Precision (mAP).
Intersection over Union (IoU) is evaluated per class for the BEV segmentation task.
We provide the results for both tasks separately to quantify the effect of temporal stereo encoding on detecting dynamic objects and static elements.

\subsection{Baseline and Implementation}\label{sec:baseline}
For uniform comparison, we set BEVFormer \cite{li2022bevformer} as the baseline model and our implementation follows the published code from both BEVFormer \cite{li2022bevformer} and UniAD \cite{hu2023uniad}.
The default BEVFormer implementation comprises 6 transformer encoder layers, each composed of self-attention, cross-attention, and feed-forward operations, with layer normalization applied after each steps.
Temporal aggregation is facilitated by the self-attention mechanism, which combines information from the previous BEV grid and queries in a recurrent refinement.
We adopt the training configuration of BEVFormer \cite{li2022bevformer} with 24 epochs and batch size 1.
The model training is performed in parallel on 8/16 NVIDIA A100 GPUs.
AdamW is used for optimization with a learning rate of $2 \times 10^{-4}$.
The size of the BEV grid is $200 \times 200$ with perception range $[ \SI{-51.2}{\meter}, \SI{51.2}{\meter} ]$ in each BEV direction, resulting in a grid cell size of $\SI{0.512}{\meter}$.

\subsection{Comparative Study}\label{sec:comparative_study}
To address the lack of direct comparison of temporal aggregation operators in literature, we conduct comparative experiments. 
For the lifting mechanism of BEV models, this has been done by Harley \etal, showing with SimpleBEV \cite{harley2023simplebev} that most performance differences come not from the mechanism itself but from other hyperparameter changes such as batch size.
In this study, we evaluate various temporal aggregation operators: temporal self-attention, max pooling, and convolution.
We replace the self-attention mechanism in our baseline model with the alternative temporal aggregation operators while keeping the remaining architecture constant.
We also evaluate the effect of the time horizon on temporal self-attention.
All experiments are trained from scratch on the full NuScenes dataset to be most representative.

{\begin{table*}[tb]
    \vspace{0.4em}
	\caption{Quantitative results of comparative study (first section) and TempBEV model (second section) on NuScenes \textit{val} dataset. Results reported for 3D object detection and BEV segmentation learning task. Aggregation is parallel (P) or recurrent (R). Results improving on the BEVFormer baseline are marked in bold, the best results are underlined.}

	\label{tab:quantitative_results_map_seg}
	\centering

 \resizebox{\textwidth}{!}{

 \begin{NiceTabular}{l|ccc|cc|r|cc|ccc}
    \toprule
    
    \multicolumn{1}{l}{\textbf{Methods}}& \multicolumn{3}{c}{\textbf{Aggregation}}&  \multicolumn{2}{c}{\textbf{Feature Space}}& \multicolumn{1}{r}{\textbf{\# Params}}&\multicolumn{2}{c}{\textbf{3D Detection} [\%]} &\multicolumn{3}{c}{\textbf{BEV Seg. IoU} [\%]} \\

    Aggregation Operator                   &  R &P & \# Steps &  Image & BEV   &[Mio.]  & NDS   & mAP   & Road & Lane & Cross \\ 
    \midrule
    No temporal aggregation                &         && 1   &  &\cmark &96.31 & 43.78  & 36.35 & 75.14 & 37.64    & 22.05    \\
    Attention (BEVFormer) - Baseline       &         \cmark && 4   &  &\cmark &97.69  & 50.25  & 39.82 & 75.65 & 38.22  & 23.89      \\
    Max pooling                            &         \cmark && 4   &  &\cmark &96.31  &41.74     &  34.12     &   73.84   &  36.09       &    19.98       \\
    Convolution 1x1&         \cmark && 4   &  &\cmark &97.10  & 47.42      & 37.95   & 74.72     &  37.05  & 22.57  \\
    Convolution 3x3&         \cmark && 4   &  &\cmark &103.39 & 48.36     & 38.48      & 75.37     & 37.72        &  23.74        \\ 
    Convolution 5x5&         \cmark && 4   &  &\cmark &115.98 & 48.95     & 39.61    & \textbf{76.51}     & \textbf{39.40}      & \textbf{25.64}      \\ 
    Temporal Stereo Encoder (256C, pret.)                &  &\cmark & 2           &         \cmark & & 101.49 &  44.23    &  36.47     &  \textbf{75.74}    &   38.21      &  23.17       \\ \midrule
    TempBEV (64C, single resolution)  &  \cmark &\cmark & 4 / 2 &  \cmark &\cmark &98.67  & \textbf{50.44} & \textbf{40.56} & \textbf{76.04} & \textbf{38.74} & \textbf{24.49}      \\
    TempBEV (64C)                     &  \cmark &\cmark & 4 / 2 &  \cmark &\cmark &99.55  &\textbf{50.76} & 39.80 & \textbf{76.30} & \textbf{38.37} & 23.57      \\
    TempBEV (64C, camera-specific)    &  \cmark &\cmark & 4 / 2 &  \cmark &\cmark &101.48 &\textbf{50.62} & \textbf{40.21} & \textbf{75.79} & \textbf{38.27} & 23.56      \\ 
    TempBEV (256C)                    &  \cmark &\cmark & 4 / 2 &  \cmark &\cmark  &102.87 &\textbf{50.84}   & \textbf{40.84}     & \textbf{76.76}    & \underline{\textbf{39.53}}        & \underline{\textbf{25.82}}       \\ 
    \textbf{TempBEV (256C, pretrained)} &  \cmark &\cmark & 4 / 2 &  \cmark &\cmark &102.87 & \underline{\textbf{51.31}}  &  \textbf{41.26}    &  \underline{\textbf{76.85}}   &  \textbf{39.34}       &  \textbf{25.74}      \\
    TempBEV (1024C, pretrained)       &  \cmark &\cmark & 4 / 2 &  \cmark &\cmark &121.75 &\textbf{51.28}     &  \underline{\textbf{41.52}}     & \textbf{76.35}     & \textbf{39.43}        &   \textbf{25.30}        \\
    \bottomrule

    \end{NiceTabular}
    }
    \vspace{-0.6em}
\end{table*}}

Table \ref{tab:quantitative_results_map_seg} shows the quantitative results of different temporal aggregation methods, the first part covers the comparative study.
Similar to Table \ref{tab:temp_bev_survey}, we list aggregation operator, recurrent or parallel aggregation, number of time steps, and feature space.
Additionally, the model size is shown by the number of learned parameters.
All models are trained on 3D object detection and BEV segmentation in a multi-task learning setup.
Results for both tasks are reported, going beyond most works that only report results on object detection.

\textit{Attention (BEVFormer)} is the baseline implementation of the BEVFormer model.
The size of the model is \SI{97.69}{\million} parameters, consisting of the encoder and object detection head as well as the BEV segmentation head (taken from from UniAD \cite{hu2023uniad}).
Our NDS and mAP scores are roughly on par with the reported values in the BEVFormer paper \cite{li2022bevformer} (NDS \SI{50.3}{\percent} \vs \SI{52.0}{\percent}, mAP \SI{39.8}{\percent} \vs \SI{41.2}{\percent}).
Also, our IoU scores are in the same range (Road IoU \SI{75.7}{\percent} \vs \SI{77.5}{\percent}, Lane IoU \SI{38.2}{\percent} \vs \SI{23.9}{\percent}).
Results slightly differ because our implementation of the BEV segmentation task excludes classes that represent dynamic objects.
To quantify the benefit of temporal aggregation, we train the baseline with no temporal aggregation by removing the temporal self-attention mechanism.
This shrinks the model size from \SI{97.69}{\million} to \SI{96.31}{\million} parameters.
The performance drop on object detection tasks is very strong and roughly matches what is reported in the BEVFormer paper (\SI{-6.5}{} percentage points (\SI{}{pts}) \vs \SI{-7.2}{pts} in NDS).
We additionally report the BEV segmentation values, which reveal that the temporal aggregation of BEVFormer only has a small benefit for segmenting static elements (\eg, Road IoU only drops from \SI{75.65}{\percent} to \SI{75.14}{\percent}).

Max pooling is used as a simple, parameter-free approach (\SI{96.3}{\million} parameters, equal to BEVFormer with no temporal aggregation) to assess the relevance of learnable weights for temporal aggregation.
In this study, we utilize top-k max pooling \cite{liu2015convolutional} with $k=256$ to map the 512-channel concatenation of  $X_{t-1}$ and queries $Q$ to 256 channels.
The results of max pooling are worse than the model with no temporal aggregation.
We assume the reason is that max pooling cannot separate the inputs from times $t-1$ and $t$, and mixes features, resulting in reduced performance.

\begin{figure*}[tpb]
	\centering
    \vspace{0.4em}
	\includegraphics[width=\textwidth]{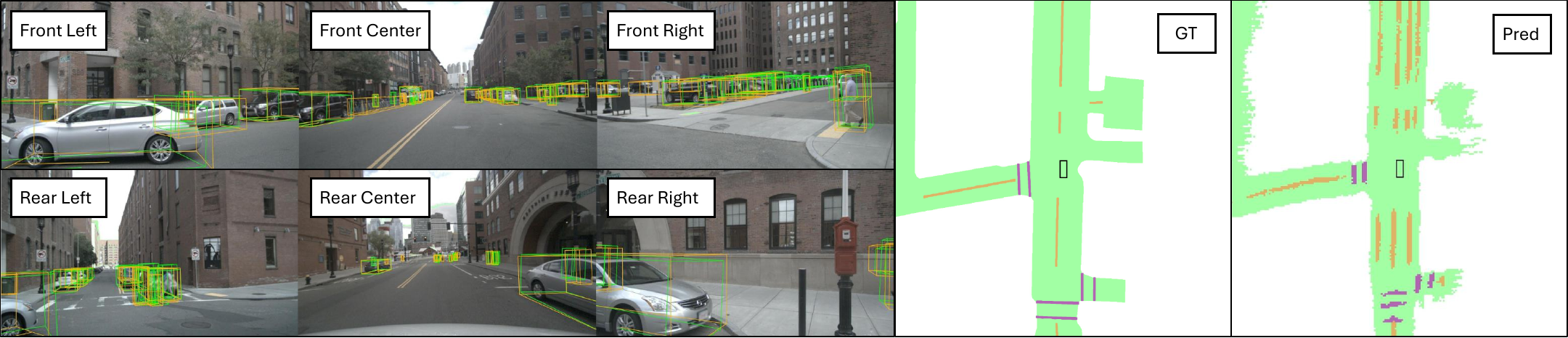}
	\caption{Qualitative result of TempBEV model on 3D object detection shown on the left (green: prediction, orange: GT) and BEV segmentation shown on the right (green: road, orange: lane, purple: crossing, black: ego vehicle).}
    \vspace{-0.3em}
	\label{fig:tempbev_qualitative_result}
 \end{figure*}

As indicated in the survey, convolution is another widely employed mechanism for temporal aggregation. 
In our comparative experiments, we utilize one simple convolution layer with 512 input channels and 256 output channels to aggregate information from $X_{t-1}$ and $Q$.
We use kernel sizes 1x1, 3x3, and 5x5 to evaluate the impact of local context on aggregation performance.
As expected, the object detection performance of the simplest convolution operation, \ie, 1x1 (NDS \SI{47.42}{\percent}), is between no temporal aggregation (NDS \SI{43.78}{\percent}) and attention (NDS \SI{50.25}{\percent}).
When adding spatial context with a 3x3 or 5x5 kernel, the metrics further increase (NDS \SI{48.36}{\percent} and \SI{48.95}{\percent})
However, bigger kernel sizes also increase the number of parameters, with \SI{97.10}{\million} for 1x1, \SI{103.39}{\million} for 3x3, and \SI{115.99}{\million} for 5x5.
Overall, compared to the more sophisticated temporal self-attention operator, a basic 1-layer convolution captures a substantial fraction of the benefits of aggregating temporal information. 

\textit{Flow (256D, pretrained)} is the ablated TempBEV model and reports the performance of flow-based image space aggregation without the recurrent BEV space aggregation.
The results show slight improvements on all metrics over no temporal aggregation, indicating limited effectiveness of flow-based parallel aggregation in image space by itself.

\subsection{Quantitative Results of TempBEV Model}\label{sec:quantitative_results}
The second part of Table \ref{tab:quantitative_results_map_seg} shows the quantitative results of different variants of the TempBEV model, serving as an ablation study of the architecture. 
TempBEV is characterized by the hidden size of its temporal stereo encoder, which is the number of channels of the latent space created by $f_{\operatorname{img}}$.
Already in its simplest form, the TempBEV model with a hidden size of 64 (64C) gives an improvement over the BEVFormer baseline with a similar NDS and improved mAP (\SI{40.56}{\percent} \vs \SI{39.82}{\percent}).
When combining the temporal stereo features with the intermediate camera features on all resolutions in the parallel aggregation module, the NDS increases to \SI{50.76}{\percent}.
This however comes with reductions in mAP, Lane IoU, and Pedestrian Crossing IoU.
We do not choose the single resolution approach since we assume that combining features in multiple resolutions could be beneficial for bigger hidden sizes.
The temporal stereo encoder operates on raw images, which have very different flow patterns based on the camera mounting position.
One idea is that camera-specific instances of the temporal stereo encoder could better learn the particular flow patterns of each camera mounting position (\eg, flow for the front camera moves away from the center, but towards the center for the rear camera).
The results are mixed with a small benefit only on mAP (increasing to \SI{40.21}{\percent} from \SI{39.80}{\percent}).
Since the number of parameters increases from \SI{99.55}{\million} to \SI{101.48}{\million}, we stick with one shared temporal stereo encoder.
We assume that the NuScenes dataset is not large enough to provide sufficient guidance to train camera-specific temporal stereo encoders individually.
For this reason, we evaluate transfer learning by using a temporal stereo encoder pretrained on optical flow datasets.
We use the encoder of a FlowNetS, the simple variant of FlowNet \cite{dosovitskiy2015flownet}, that is pretrained on Flying Chair Dataset \cite{dosovitskiy2015flownet}.
On a FlowNet with a hidden size of 256, we see a strong improvement in object detection (NDS \SI{51.31}{\percent} \vs \SI{50.84}{\percent}).
The BEV segmentation results remain similar, indicating that transfer learning from optical flow estimation is primarily helpful in detecting dynamic objects in the scene.
The hidden size of the FlowNet encoder is ablated.
Increasing the hidden size from 64 to 256 without pretraining shows a strong improvement on all metrics with the number of parameters slightly increasing from \SI{99.55}{\million} to \SI{102.87}{\million}
Further increasing hidden size to 1024 for FlowNet with pretraining increases the number of parameters drastically to \SI{121.75}{\million}, but shows no additional improvement (NDS \SI{51.31}{\percent} \vs \SI{51.28}{\percent}).
Hence, for our final model we define TempBEV using a pretrained optical flow encoder with hidden size 256.
This adds \SI{6.18}{\million} parameters (\SI{+5.3}{\percent}) to the BEVFormer baseline and changes inference speed on an Nvidia A6000 GPU from 2.18 FPS to 1.94 FPS (\SI{-11.0}{\percent}).

With TempBEV, we combine temporal aggregation in image and BEV space to test our hypothesis from Section \ref{sec:approach_temporal_aggregation}.
Compared to the BEVFormer baseline model, even with the simplest optical flow encoder from FlowNet, TempBEV performs better by \SI{+1.06}{pts} NDS, \SI{+1.44}{pts} mAP, \SI{+1.20}{pts} Road IoU, \SI{+1.12}{pts} Lane IoU, and \SI{+1.85}{pts} Pedestrian Crossing IoU.
This confirms our hypothesis of the complementary nature of image and BEV space temporal aggregation.
Starting from the baseline with no temporal aggregation and adding the image space aggregation improves NDS by just \SI{+0.45}{pts}.
Starting from the baseline with BEV space aggregation and adding image space aggregation improves NDS even more by \SI{+1.06}{pts}.
This highlights that image and BEV space aggregation are not just complementary but also show synergy effects.
The results show the effectiveness of our adjustments to the temporal aggregation mechanism.
We leave it for future work to use these insights to improve the latest state of the art models.

\subsection{Qualitative Results}\label{sec:qualitative_results}
Fig. \ref{fig:tempbev_qualitative_result} shows qualitative results of TempBEV on 3D Object Detection and BEV Segmentation tasks.
Frames from 6 cameras are overlaid with ground truth (GT) and prediction bounding boxes projected into the camera view. 
On the right, GT and predicted BEV segmentation are visualized.
TempBEV accurately predicts the bounding boxes of dynamic objects in 3D and the BEV segmentation of the road class.
It predicts dividers (class lane) that separate the bus lane behind the ego vehicle, that are missing in GT.

\section{Conclusion} \label{sec:conclusion}
In this paper, we presented a survey and a comparative study on temporal aggregation mechanisms for BEV encoders.
Based on the gained insights, we proposed TempBEV, a novel BEV model that combines recurrent aggregation in BEV space with parallel aggregation in image space.
An optical flow encoder is used to extract temporal stereo features directly from pairs of subsequent camera frames in image space.
Experiments show that TempBEV provides a significant increase in performance by using a simple optical flow encoder, with further potential to be expected from more sophisticated approaches.
The results suggest a synergy effect between temporal aggregation in different representations, making a strong case for combined aggregation in both image space and BEV space.
We validated our hypothesis based on the BEVFormer model, which is commonly used for BEV encoding \cite{hu2023uniad, yuan2024streammapnet}, and expect similar synergy effects when applied to other state of the art BEV encoders.

\vspace{0.4em}

\bibliography{Literature}{}
\bibliographystyle{IEEEtran}

\end{document}